\documentclass[letterpaper, 10 pt, conference]{ieeeconf}  

\IEEEoverridecommandlockouts                              

\usepackage{amssymb}  
\usepackage{graphicx}
\usepackage{multirow}
\usepackage{hyperref}
\title{\LARGE \bf
Ladder Fine-tuning approach for SAM integrating complementary network
}

\author{Shurong Chai$^{1}$, Rahul Kumar Jain$^{1}$, Shiyu Teng$^{1}$, Jiaqing Liu$^{1}$, Yinhao Li$^{1}$,  Tomoko Tateyama$^{2}$ and *Yen-wei Chen$^{1}$
\thanks{$^{1}$ Shurong Chai, Rahul Kumar Jain, Shiyu Teng, Yinhao Li, Jiaqing Liu and Yen-wei Chen are with the College of Information Science and Engineering, Ritsumeikan  University, Shiga, Japan. E-mail: is0538kr@ritsumei.ac.jp}
\thanks{$^{2}$Tomoko Tateyama is with Department of Intelligent Information Engineering, Fujita Health University, Japan}%
\thanks{*Corresponding Authors:Yen-Wei Chen (chen@is.ritsumei.ac.jp)
}%
}

\begin{document}

\maketitle
\thispagestyle{empty}
\pagestyle{empty}

\begin{abstract}

Recently, foundation models have been introduced demonstrating various tasks in the field of computer vision. These models such as Segment Anything Model (SAM) are generalized models trained using huge datasets. Currently, ongoing research focuses on exploring the effective utilization of these generalized models for specific domains, such as medical imaging. However, in medical imaging, the lack of training samples due to privacy concerns and other factors presents a major challenge for applying these generalized models to medical image segmentation task. To address this issue, the effective fine tuning of these models is crucial to ensure their optimal utilization. In this study, we propose to combine a complementary Convolutional Neural Network (CNN) along with the standard SAM network for medical image segmentation. To reduce the burden of fine tuning large foundation model and implement cost-efficient trainnig scheme, we focus only on fine-tuning the additional CNN network and SAM decoder part. This strategy significantly reduces trainnig time and achieves competitive results on publicly available  dataset. The code is available at 
\href{https://github.com/11yxk/SAM-LST}{https://github.com/11yxk/SAM-LST}.

\end{abstract}

\section{INTRODUCTION}

Medical image segmentation plays a crucial role in healthcare. It aims to segment various body organs including the liver, brain, and lesions, using various medical imaging modality such as  X-rays, CT scans, MRI scans, or ultrasound images. Hence, it significantly aids clinicians in diagnosing, treatment planning and post-treatment monitoring. Over the past decade, Convolution Neural Network (CNN) has become popular for a wide range of computer vision tasks.

Recently, Long et al. \cite{long2015fully} proposed the Fully Convolutional Network (FCN). This approach enables the processing of input images of any size and generates segmentation results by replacing fully connected layers with convolutional layers. U-Net \cite{ronneberger2015u}, developed by Ronneberger et al., is the most widely used architecture for medical image segmentation. It includes an encoder and a decoder with skip connections between the corresponding layers to preserve important features. The encoder pathway down-samples the input image while capturing high-level features. Whereas the decoder pathway performs up-sampling of feature maps to predict the segmentation results. Zhou et al. \cite{zhou2018unet++} extends the U-Net architecture by introducing a nested skip connections scheme, which allows to capture multi-scale contextual information and better integration of features from different levels. Chen et al. \cite{chen2017deeplab} proposed the Deeplab series models, which includes the concept of atrous/dilated convolution operations and fully connected conditional random fields. 

Recently, Transformer \cite{vaswani2017attention} have been introduced in the Computer Vision (CV) field, which was originally designed for Natural Language Processing (NLP). Transformers can capture the long-range dependencies in comparison to the conventional CNN architecture. Dosovitskiy et al. \cite{dosovitskiy2020image} proposed the Vision Transformer (ViT) for image classification employing self-attention mechanism. Following, Chen et al. \cite{chen2021transunet} presented the TransUNet, which employs the ViT for segmentation task. TransUNet jointly utilized CNN and ViT to obtain the local and global contextual features from input images. Tang et al. \cite{tang2022self} presented the Swin UNETR that employs ViT model as main encoder for feature extraction. Zhou et al. \cite{zhou2021nnformer} proposed a pure Transformer framework, which uses ViT in both encoder and decoder parts. Cao et al. \cite{cao2021swin} proposed the Swin-UNet that adopts the Swin-transformer \cite{liu2021swin} architecture for segmentation task. 

Nowadays, foundation model \cite{bommasani2021opportunities} has demonstrated its capability in the field of NLP. Currently, Segment Anything Model (SAM) \cite{kirillov2023segment} is introduced for various Computer Vision tasks. In SAM, the concept of prompt learning using foundation model enables multiple tasks on unseen images. It allows zero-shot transfer to various tasks through effective prompt engineering. While applying the SAM model directly to a domain-specific task, such as medical image segmentation, normally does not yield satisfactory performance. Despite SAM is trained using more than 11M images with 1 billion ground truth masks, its application for medical image segmentation poses challenges due to the distinct characteristics of medical images compared to real-world images. Additionally, the scarcity of medical data is also a major issue in fine tuning SAM. Therefore, it is essential to effectively fine-tune the SAM for medical image datasets. 

Nowadays, various fine-tuning methods have been introduced to optimize the SAM on different domains \cite{wu2023medical, zhang2023customized, chen2023sam}. Some approaches perform adapter-based fine tuning for the SAM network.  However, these adapter-based methods often require significant effort and resource costs for training the model. Different form previous studies, our work introduces a novel approach by combining an additional CNN as a complementary Encoder within SAM architecture. Our approach draws inspiration from the Ladder-Side Tuning (LST) network for Transformers \cite{sung2022lst}. Our proposed approach enables the flexible integration of an additional network while avoiding backpropagation on the entire large model (i.e., SAM Encoder), leading to faster training and reduced resource costs. The additional CNN network can be easily replaced by other designs, including Transformers, based on the specific task requirements. We incorporate a pretrained ResNet18 \cite{he2016deep} as an additional network. During training, only the parameters of the additional CNN and Decoder parts are fine-tuned while keeping the original SAM encoder parameters frozen. Our contribution can be summarized as follows:

(1) We propose to combine an additional CNN for fine tuning the SAM specially on medical image segmentation task. 

(2) The proposed approach offers flexibility in designing an additional network while minimizing resource costs by avoiding backpropagation on the entire model. 

(3) Our method achieves competitive results compared to state-of-the-art methods on a publicly available multi-organ segmentation dataset, without the use of any prompts.

\section{RELATED WORK}
\subsection{Medical image segmentation}
Accurate and reliable medical image segmentation is essential in assisting medical diagnosis. Over the past few years, numerous segmentation approaches have been put forward. CNN-based networks in particular have shown remarkable success in this task \cite{ronneberger2015u,zhou2018unet++,chen2017deeplab}. Recently, some advanced Transformer-based networks are also proposed \cite{tang2022self,zhou2021nnformer,cao2021swin}, achieving a new milestone in this task. Although the significant progress in medical image segmentation, it remains a challenging task due to factors such as limited data availability and requirement of clinical experts to annotate the data. These factors often result in poor generalization of models.
\subsection{Foundation models}
Foundation models refer to a model that is trained on broad data and can be adapted to a wide range of downstream tasks \cite{bommasani2021opportunities}. This paradigm usually contains some other techniques such as self-supervised learning, transfer learning and prompt learning. An example of a foundation model is the Generative Pre-trained Transformer (GPT) series, these models are pre-trained on vast amounts of text data from various sources. These models have been significantly contributing to the advancement of Natural Language Processing (NLP). Specifically, GPT-3 \cite{brown2020language}, one of the Large Language Models (LLMs), with 175 billion parameters, can be applied to wide range of tasks including translation, question-answering, and cloze tasks. Another notable work is Contrastive Language-Image Pre-Training (CLIP) \cite{radford2021learning}, which employed a large-scale dataset with paired images and their corresponding textual descriptions. CLIP can effectively retrieve images according to given text prompts, which have many applications such as image classification and image generation. These foundation models have achieved state-of-the-art performance. These models have immense future directions for advancement in various fields. 

\subsection{Parameter-Efficient Fine-Tuning}
Although foundation models have made significant achievements, they still face some limitations, such as requirement of abundant labeled data for training and substantial computational resources due to their huge number of parameters. To reduce the large computational cost, Parameter-Efficient Fine-Tuning (PEFT) is introduced by training a small set of parameters of the existing models or trainnig of newly added parameters in the architecture \cite{lialin2023scaling}. Houlsby et al. \cite{houlsby2019parameter} proposed to add a small subnetwork, called an “adapter” into the original foundation model. Lester et al. \cite{lester2021power} proposed to prepend a trainable Tensor into original model input. Sung et al. \cite{sung2022lst} introduced a novel Ladder-Side Tuning (LST) paradigm, which only finetunes a small Transformers network incorporated beside the original model. In this architecture design, the parameters of the newly incorporated networks are only updated to save the computation cost. Ben-Zaken et al. \cite{zaken2021bitfit} proposed to train only the biases of original network, which is also a simple and effective approach. In general, PEFT-based methods are GPU-friendly, allowing to employ foundation model for various downstream tasks even with limited computational resources.
\section{METHOD}

\begin{figure*}[h]
\begin{center}
\includegraphics[width=1\linewidth]{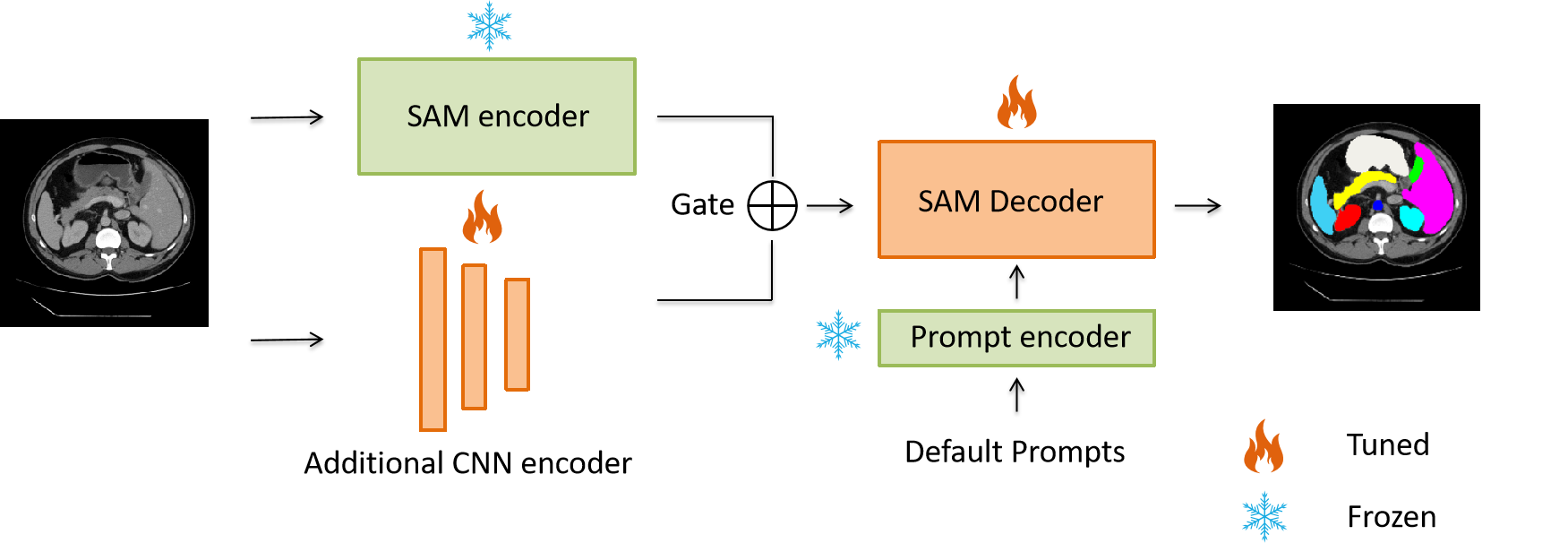}
\caption{Overview of our proposed method.}
\label{fig:1}
\end{center}
\end{figure*}

\subsection{Segment Anything Model}

The Segment Anything Model (SAM) \cite{kirillov2023segment} is the first attempt of foundation models in segmentation task. SAM consists of three components, these are image encoder, prompt encoder and mask decoder. The image encoder employs an MAE \cite{he2022masked} pre-trained ViT network \cite{dosovitskiy2020image} to extract image features. The prompt encoder enables four types of prompt inputs: points, boxes, text and masks.  The points and boxes are embedded with positional encoding \cite{tancik2020fourier} while the text is embedded with text encoder from CLIP \cite{radford2021learning}. Masks are embedded using convolution operations. The mask decoder is designed to map the image embedding and prompt embedding in a lightweight manner. These two types of embeddings interact using cross-attention module, using one embedding as query and another embedding as key and value vectors. Finally, the Transposed convolutions are used to up-sample the features. The mask decoder has the capability to generate multiple results as the provided prompts might have ambiguity. The default number of outputs is set to three. It is worth to mention that the image encoder extracts image features only once for each input  image. After that the lightweight prompt encoder and mask decoder can interact with users based on different input prompts in a web browser in real-time. The SAM is trained using more than 11M images and 1B masks. The experimental results demonstrate the superior zero-shot transfer ability. As implied by its name, the model can almost segment anything, even in cases that have not seen before (unseen test samples).

\begin{figure}[h]
\begin{center}
\includegraphics[width=0.5\linewidth]{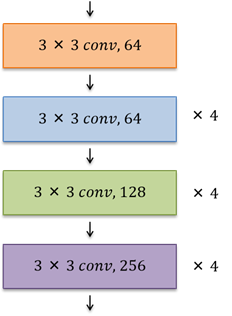}
\caption{The architecture of CNN Encoder. In this figure, we omit the activation function, batch normalization layer and residual connections for simplicity.}
\label{fig:2}
\end{center}
\end{figure}

\subsection{Ladder-Side Tuning with SAM}
Here we describe our proposed scheme and integrated network. The overview is shown in Figure 1. To effectively apply the SAM model for medical image segmentation task, we propose to add a lightweight side network while avoiding the backpropagation through entire SAM model. We only update the parameters of the SAM Decoder and integrated CNN network to finetune the SAM on medical dataset. Formally, given an input sample x $x \in \mathbb{R}^{H \times W \times C}$ where $H$, $W$, $C$ denote height, width, channel numbers. The input image is first fed into SAM image encoder and CNN encoder simultaneously:

\begin{equation} 
x_{sam}=SAM\_encoder(x) \in \mathbb{R}^{H^{\prime} \times W^{\prime} \times C^{\prime}}
\end{equation}

\begin{equation} 
x_{cnn}=CNN\_encoder(x) \in \mathbb{R}^{H^{\prime} \times W^{\prime} \times C^{\prime}}
\end{equation}

The architecture of CNN encoder is shown in Figure.2. The design follows the ResNet18 \cite{he2016deep}, which includes skip connections. However, the network is modified to generate feature maps size of the same size as SAM encoder. To accomplish this, we only use 13 layers instead of using all 18 layers (ResNet18 contains 18 convolutional layers). To determine the importance of features extracted from two networks, i.e., SAM and integrated CNN, we propose to incorporate a learnable gate (weight parameter) $\alpha$  when combining these two extracted feature maps:

\begin{equation} 
x=\alpha \cdot x_{sam}+(1-\alpha) \cdot x_{cnn}
\end{equation}

\subsection{Loss functions}
The combination of Cross Entropy loss and Dice loss is used to fine tune the network. 

\begin{equation} 
L=(1-\lambda) \cdot L_{Cross\_entropy}+\lambda \cdot L_{Dice}
\end{equation}
Where $\lambda$ is a hyperparameter and its value is set to 0.8 based on our experiments.

\begin{table*}[]
\begin{center}
\caption{Comparison with State-of-the-arts}
\begin{tabular}{l|c|c|c|c|c|c|c|c|c|c}
\hline
Method              & DSC(\%)$\uparrow$ & HD95(mm)$\downarrow$ & Aorta & Gallbladdr & Kidney(L) & Kidney(R) & Liver & Pancreas & Spleen & Stomach  \\ \hline\hline
V-Net{\cite{milletari2016v}}       & 68.81   & -        & 75.34 & 51.87      & 77.10     & 80.75     & 87.84 & 40.05    & 80.56  & 56.98        \\
DARR{\cite{fu2020domain}}        & 69.77   & -        & 74.74 & 53.77      & 72.31     & 73.24     & 94.08 & 54.18    & 89.90  & 45.96         \\
R50 U-Net{\cite{chen2021transunet}}    & 74.68   & 36.87    & 87.74 & 63.66      & 80.60     & 78.19     & 93.74 & 56.90    & 85.87  & 74.16         \\
U-Net{\cite{ronneberger2015u}}        & 76.85   & 39.70    & 89.07 & 69.72      & 77.77     & 68.60     & 93.43 & 53.98    & 86.67  & 75.58        \\
R50 Att-UNet{\cite{chen2021transunet}} & 75.57   & 36.97    & 55.92 & 63.91      & 79.20     & 72.71     & 93.56 & 49.37    & 87.19  & 74.95        \\
R50 ViT{\cite{chen2021transunet}}      & 71.29   & 32.87    & 73.73 & 55.13      & 75.80     & 72.20     & 91.51 & 45.99    & 81.99  & 73.95        \\
TransUnet{\cite{chen2021transunet}}    & 77.48   & 31.69    & 87.23 & 63.13      & 81.87     & 77.02     & 94.08 & 55.86    & 85.08  & 75.62     \\
MTM{\cite{wang2022mixed}}         & 78.59   & 26.56    & 87.92 & 64.99      & 81.47     & 77.29     & 93.06 & 59.46    & 87.75  & 76.81    \\
SwinUNet{\cite{cao2021swin}}    & 79.12   & 21.55    & 85.47 & 66.53      & 83.28     & 79.61     & 94.29 & 56.58    & 90.66  & 76.60     \\
SAMed{\cite{zhang2023customized}}  & 81.88   & 20.64    & 87.77 & 69.11     & 80.45     & 79.95    & 94.80 & 72.17   & 88.72  & 82.06     \\ 
MISSFormer{\cite{huang2021missformer}}  & 81.96   & 18.20    & 86.99 & 68.65      & 85.21     & 82.00     & 94.41 & 65.67    & 91.92  & 80.81      \\ \hline
Ours                & 79.45   & 35.35    & 88.05 & 66.53      & 81.45    & 75.69     & 94.56 & 63.08   & 87.71  & 78.51    \\ \hline
\end{tabular}
\end{center}
\end{table*}

\section{EXPERIMENTS}
\subsection{Dataset} 
We use Synapse dataset for evaluation, which is a publicly available multi-organ segmentation dataset of the MICCAI 2015 Multi-Atlas Abdomen Labeling Challenge. It includes 30 abdominal CT scans. Following the previous work \cite{chen2021transunet}, a total number of 18 cases are used for training and 12 cases are used for test. We report the results in terms of Dice Similarity Coefficient (DSC) and 95\% Hausdorff Distance (HD95) on 8 abdominal organs (i.e., aorta, gallbladder, spleen, left kidney, right kidney, liver, pancreas, spleen, stomach).

\subsection{Implementation detail}
The input image resolution is set to 224×224. We use random rotation and flipping operations for data augmentation following the previous work \cite{zhang2023customized}. We use ViT-B SAM model as foundation backbone model. We do not fine tune the SAM encoder and prompt encoder. While we fine tune only the ‘output\_upscaling’ part of SAM decoder to avoid overfitting. The integrated CNN encoder is pretrained on ImageNet, provided by PyTorch Torchvision library. The framework is trained using Adam optimizer with a batch size of 24 for 200 epochs. The learning rate is set to 0.001. The warmup strategy is applied for 250 iterations. The experiment is conducted using two RTX 3090 graphics cards.

\subsection{Experiment results}
Table.I reports experiment results and comparison with other state-of-the-arts methods. Our proposed method achieves 79.45\% DSC and 35.35mm HD95 scores. We also observe that the value of learnable weight parameter is 0.44. Our method achieves a competitive score while surpassing most of state-of-the-art methods. Some segmentation results are shown in Figure.3. However, the design of integrated CNN Encoder and learnable weight parameter can be modified to analyze and evaluate the performance of the proposed approach. We believe that utilizing a Transformer or other effective network design will yield even higher performance. In the future, we will explore advanced design choices to achieve the best results.

\subsection{Ablation study}
Our ablation study is conducted to evaluate the effectiveness of integrating the CNN encoder with the SAM encoder. The Table. II demonstrates that the SAM model achieves a Dice Score of only 1.73\% without fine-tuning for medical images. It is important to note that no prompts are used during this training and evaluation, resulting in a low score due to the direct application of a generalized model. With the fine-tuning method applied to the entire SAM, the accuracy improves to 58.97 Dice score. Similarly, when the CNN encoder is combined with the SAM decoder module, the performance remains at 78.05 Dice score. This highlights the need for effective fine-tuning methods. However, by integrating the CNN encoder with the SAM network and utilizing a learnable gate (weight parameter), the accuracy significantly improves to 79.45 Dice score. Moreover, we also observe a significant reduction in training time, with a decrease of approximately 30\% to 40\% compared to other fine-tuning methods \cite{wu2023medical, zhang2023customized, chen2023sam}. Our proposed method is much cost-effective in terms of resource utilization.

\begin{table}[]
\caption{Ablation results}
\begin{center}
\begin{tabular}{l|l|l}
\hline
Method                 & DSC(\%)$\uparrow$ & HD95(mm)$\downarrow$ \\ \hline
SAM                    & 1.73    & 260.98   \\
Ours(CNN Encoder only) & 78.05   & 29.11    \\
Ours(SAM Encoder only) & 58.97   & 101.60   \\
Ours                   & 79.45   & 35.35    \\ \hline
\end{tabular}
\end{center}
\end{table}

\begin{figure}[h]
\begin{center}
\includegraphics[width=1\linewidth]{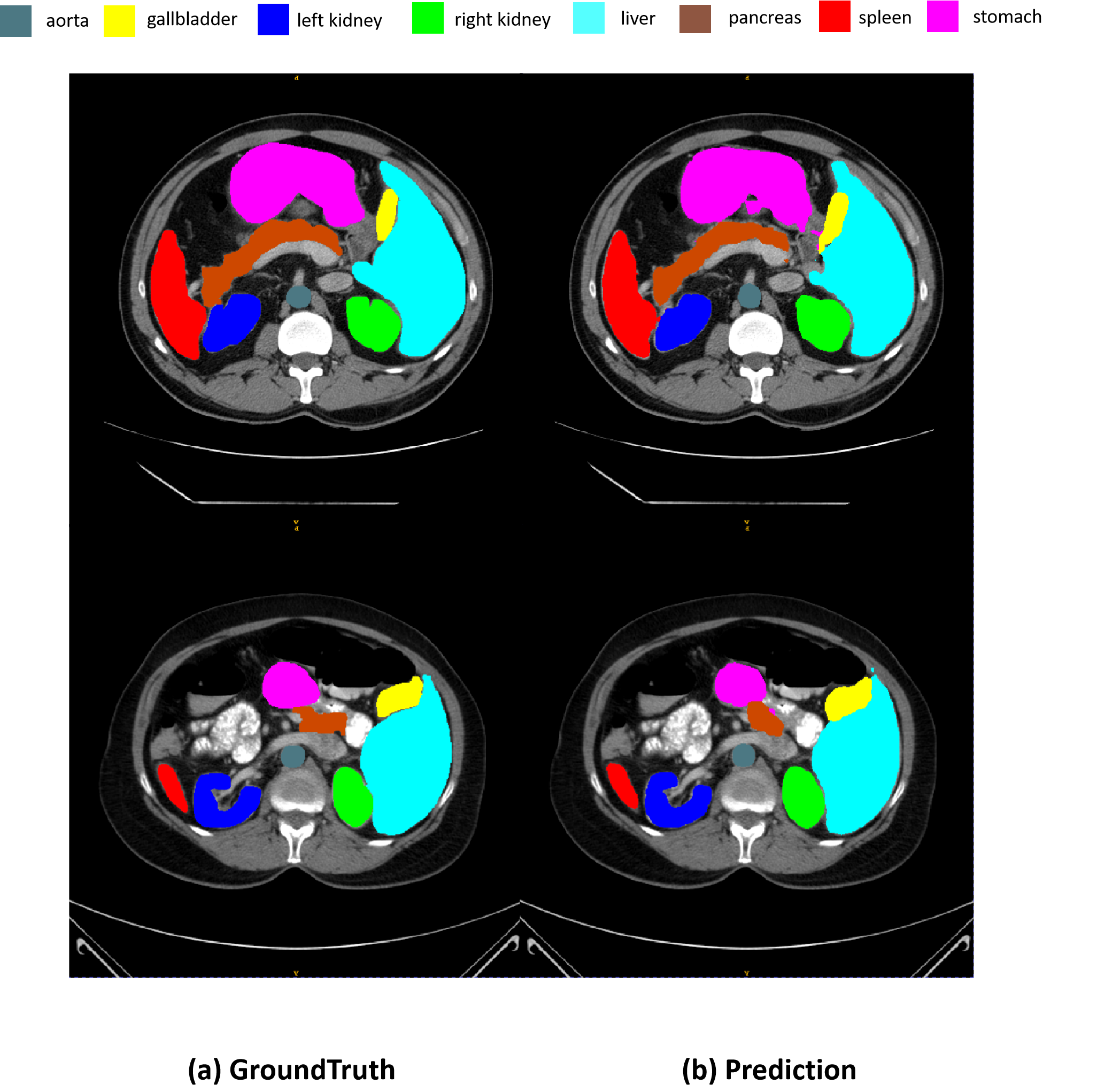}
\caption{Segmentation results on Synapse dataset.}
\label{fig:3}
\end{center}
\end{figure}

\section{CONCLUSION}

We introduce a robust and flexible fine-tuning strategy for large foundation model, specifically SAM. Our proposed approach of integrating CNN encoder while employing a learnable weight parameter achieves a significant result. This approach provides the way for new fine-tuning strategies in computer vision. Furthermore, our proposed approach minimizes resource utilization and reduces training time. In the future, we aim to explore additional fine-tuning methods to enhance performance.

\bibliographystyle{IEEEtran}


\end{document}